\documentclass[preprint,12pt]{elsarticle}
\usepackage[utf8]{inputenc}
\usepackage{graphicx}
\usepackage{indentfirst}
\usepackage{amsmath}
\usepackage{amssymb}
\usepackage{tabto}
\usepackage[table]{xcolor}
\usepackage{wrapfig}
\usepackage{hyperref}
\graphicspath{{./images/}}

\begin{document}

\begin{frontmatter}

\title{Graph Neural Networks with Trainable Adjacency Matrices for Fault Diagnosis on Multivariate Sensor Data}

\author[hse]{Alexander Kovalenko}
\author[airi,hse]{Vitaliy Pozdnyakov}
\author[airi,misis]{Ilya Makarov}

\affiliation[hse]{organization={HSE University},
            city={Moscow},
            country={Russia}}
            
\affiliation[airi]{organization={AIRI},
            city={Moscow},
            country={Russia}}

\affiliation[misis]{organization={NITU MISIS},
            city={Moscow},
            country={Russia}}

\begin{abstract}
Timely detected anomalies in the chemical technological processes, as well as the earliest detection of the cause of the fault, significantly reduce the production cost in the industrial factories. Data on the state of the technological process and the operation of production equipment are received by a large number of different sensors. To better predict the behavior of the process and equipment, it is necessary not only to consider the behavior of the signals in each sensor separately, but also to take into account their correlation and hidden relationships with each other. Graph-based data representation helps with this. The graph nodes can be represented as data from the different sensors, and the edges can display the influence of these data on each other. In this work, the possibility of applying graph neural networks to the problem of fault diagnosis in a chemical process is studied. It was proposed to construct a graph during the training of graph neural network. This allows to train models on data where the dependencies between the sensors are not known in advance. In this work, several methods for obtaining adjacency matrices were considered, as well as their quality was studied. It has also been proposed to use multiple adjacency matrices in one model. We showed state-of-the-art performance on the fault diagnosis task with the Tennessee Eastman Process dataset. The proposed graph neural networks outperformed the results of recurrent neural networks.
\end{abstract}

\begin{keyword}
Tennessee Eastman Process \sep Chemical Processes \sep Graph neural networks \sep Fault diagnosis \sep Sensor data
\end{keyword}

\end{frontmatter}

\section{Introduction}
During the production, equipment often stops due to the various faults. The process of finding the root case of the fault can take significant amount of time. The deviations of equipment parameters can lead to a defective product. It is also mean the lost of time and lost of raw materials. All this leads to financial losses for companies. To reduce the cost of production, the equipment must be in good condition and the deviations of the parameters must be corrected as soon as possible. Sometimes the causes of faults are completely non-obvious and have hidden dependencies. It creates additional diagnostic difficulties even for highly qualified experts. For decades, scientists and specialists have been developing methods to detect faulty equipment conditions and determine the types of the faults. In the literature, such problems are usually called fault detection and diagnosis (FDD).

The latest scientific discoveries and developments in the fields of electronics and computer science provide us with new opportunities in various areas of our life, including industry. New types of automation sensors have been available and computing capacity has increased. It is enabling the application of machine learning models in practice. The concept of Industry 4.0 focuses on interconnectivity, automation, machine learning, and real-time data in manufacturing \cite{ind4}. Modern and modernized production equipment consists of hundreds of sensors, data from which can be used to increase the quality and productivity of production lines. 

The multivariate sensor data gives information about the status of various equipment components affecting the production process. These data are usually presented in the form of time series. Modern machine learning approaches and classical statistical FDD methods work only with numerical values of time series and do not take into account the information about possible correlations between them. Production equipment consists of many devices that work together and deviations in the operation of one of them can affect the operation of the others. All these changes are also displayed in the signals from the sensors of these devices. The received signals can have hidden dependencies and can correlate with each other. To indicate this correlation, such data can be represented as a graph. As nodes of the graph, there can be values of time series from the sensors, and as edges, their influence on each other.

The main goal of this work was to investigate the possibility of applying graph neural networks (GNN) to industrial equipment data. These types of neural networks show good results when working with graph-structured data. The relationship between the nodes of the graph are usually described by adjacency matrices. These data are very important but not always known and can be of different quality. Over the past year, very few papers have been published on the use of GNN in fault diagnosis problems \cite{GNNFD, GNNFD2}. In these papers, the adjacency matrices are known in advance or obtained by various methods before the start of the GNN training process. In this work, options to obtain an adjacency matrix during the training process were proposed. The adjacency matrix can be created in various ways by using trainable parameters. 

One adjacency matrix can not always describe all possible relationships between sensors. Production equipment can have complex dynamic dependencies between its devices. Different modes of operation can be described by different adjacency matrices. A novel idea was proposed to learn in parallel and use several adjacency matrices during the training and evaluation processes.

All GNN models were trained on the popular FDD benchmark to determine the type of fault. The results were compared with baselines such as multilayer perceptron (MLP), 1d convolutional neural network (1DCNN) and Gated Recurrent Units (GRU).

There are three core contributions of this work:
\begin{itemize}
    \item For the FDD problem, the architecture of the GNN model with various ways to obtain a weighted adjacency matrix was investigated. It was also proposed to train and use an adjacency matrix having both positive and negative weights.
    \item The quality of the adjacency matrices obtained during model training was studied.
    \item A novel idea of using multiple adjacency matrices was proposed and explored.
\end{itemize}

The structure of this work is organized as follows. In Related Works, FDD methods and GNN-based models are briefly described. The architecture of the GNN model with trainable adjacency matrix is proposed in the Model Description section. Dataset Description contains information about the benchmark used. The Experiment section consists of three parts that describe the contributions listed above. In Conclusion, the results of this work are summarized.

\section{Related Works}
Fault detection and diagnosis (FDD) methods provide great benefits in reducing production costs and improving quality and productivity. The data structures used by these methods can be represented as graph structures. In the same time, great development of graph neural networks in recent years may allow them to be used as approaches for fault detection and diagnosis problems. Recent papers in the fields of FDD and GNN are reviewed below.

\subsection{Fault Detection and Diagnosis}
\begin{figure}[!ht]
    \centering
    \includegraphics[width=0.8\textwidth]{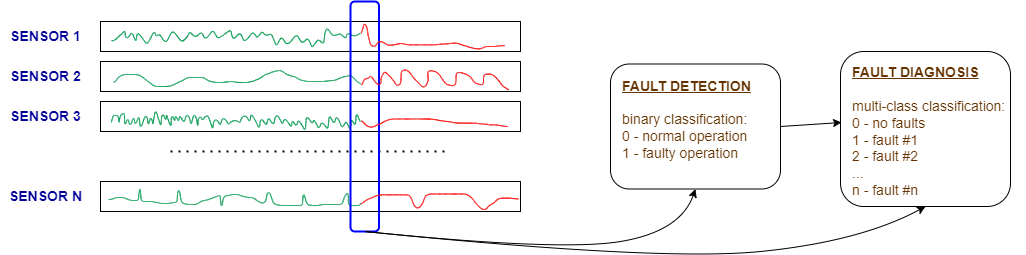}
    \caption{Fault diagnosis task can directly classify the fault or first detect abnormal state as a subtask.}
    \label{fig:fdd}
\end{figure}

Fault detection is the process of identifying that an equipment or process is not in a normal state, and fault diagnosis is the process of determining the root cause of the fault (Fig. \ref{fig:fdd}). Many different FDD techniques and approaches have been developed over the past decades \cite{review1}. These methods can be classified into data-driven \cite{datadriven1,datadriven2,datadriven3}, model-based \cite{modelbased1,modelbased2,modelbased3} and knowledge-based groups \cite{knowledgebased1,knowledgebased2,knowledgebased3}. The latter two require expert knowledge and understanding of the technological process, which does not allow to create unique FDD tools applicable to various industries. At the same time, data-driven approaches depend on the analytical models used and on the quality of historical data and could be scaled to different production/processing equipment. 

Many statistical and machine learning methods have been proposed for data-driven FDD. Dimensional reduction methods, such as principal component analysis (PCA), t-SNE, canonical variate analysis (CVA), partial least squares (PLS), represent high-dimensional sensor data in low-dimensional feature space making FDD tractable in cases of a large number of sensors \cite{pca, pls, cva}. Hotelling's $T^2$, squared prediction error (SPE), Kullback-Leibler divergence, and other statistics are calculated in feature space in order to detect faults in a process. Cheng Ji and Wei Sun reviewed statistical methods of root cause diagnosis \cite{review2}, that are divided into three groups: contribution plot-based methods \cite{contributionplotbased}, probability reasoning-based methods \cite{probareasoningbased} and causal reasoning-based methods \cite{causalreasoningbased}. Clustering methods, e.g. K-means and DBSCAN, are used as unsupervised FDD to separate sensor data into distinct groups, each of them represents some state of a process \cite{convae}. Such clusters can be manually labeled by experts for configuring a process monitoring system based on a clustering algorithm. On the other hand, supervised FDD methods, such as random forest \cite{randomforest}, support vector machine (SVM) \cite{svm}, k-nearest neighbors (k-NN) \cite{knn}, are trained to detect faults using labeled sensor data. Usually supervised methods are more accurate than unsupervised methods, but require manual labeling of each data point, which can be difficult and expensive in real industrial cases \cite{unsupervisedfdd}.

The great advances in deep learning over the past decade have affected many areas, including the field of fault detection and diagnosis. Convolutional neural networks (CNN) have become widely used in FDD tasks \cite{CNN1, CNN2}. Ildar Lomov et al. \cite{RNN} investigated the application of various architectures of recurrent and convolutional neural networks for fault diagnosis in a chemical process. Deep learning approaches show high efficiency but require a large amount of training data.

\subsection{Graph Neural Networks}
\begin{figure}[!ht]
    \centering
    \includegraphics[width=0.8\textwidth]{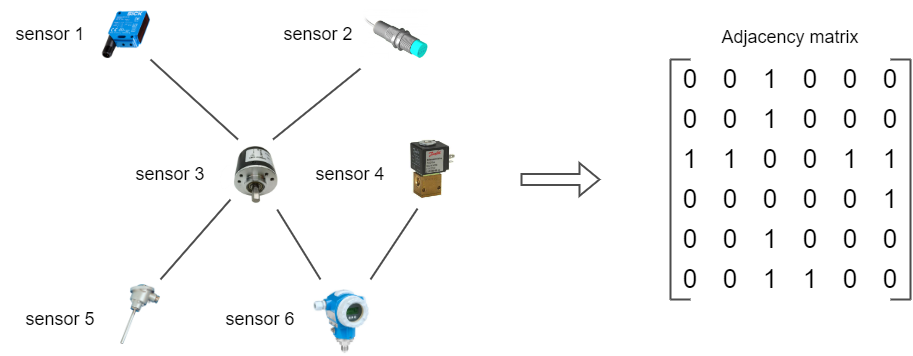}
    \caption{Relationships between devices in the form of an adjacency matrix.}
    \label{fig:adj}
\end{figure}

Multivariate sensor data can be represented in the form of graph, where the nodes are sensors. Two nodes are connected by the edge if the data from these two sensors are depending on each other. The most commonly used mathematical notation of the graph is $ G = (V, E) $, where $V$ is a set of nodes and $E$ is a set of edges. The number of nodes is denoted by $N = |V|$. The neighborhood of a node $\upsilon \in V$ is defined as $\mathcal{N} = \{u \in V|(\upsilon,u) \in E\}$, where $(\upsilon,u)$ is denoted as an edge between $\upsilon$ and $u$. The structure of a graph can be represented by adjacency matrix $A \in R^N{}^\times{}^N$ with $A_i{}_j = c > 0$ if $(\upsilon_i,u_j) \in E$ and $A_i{}_j = 0$ if $(\upsilon_i,u_j) \notin E$ (Fig. \ref{fig:adj}).

For a long time, signed directed graphs (SDG) have been widely used for failure path analysis, however, they require expert knowledge of equipment and processes \cite{SDG}. On the other hand, data-driven methods are based on only historical data. An example is representing relationships between sensors by clustering sensors into the groups and finding the correlation between them \cite{zhu2022enhanced}. Another way to represent relationships between sensors is an adjacency matrix. Graph neural networks model the dependencies between nodes on historical data using predefined adjacency matrix. The first concept of GNN was proposed in 2009 \cite{GNN}. It can be considered as a generalization of convolutional neural networks. GNNs began to develop rapidly after the publication of Kipf and Welling \cite{GCN}, in which he described graph convolutional networks (GCN) from a mathematical point of view.

The main idea behind the GCN is the aggregation of information in a node from its neighbors. Kipf et al. \cite{GCN} proposed the following layer-wise propagation rule:
\begin{equation} \label{gcn}
H^{(l+1)}=\sigma(\tilde{D}^{-\frac{1}{2}}\tilde{A}\tilde{D}^{-\frac{1}{2}}H^{(l)}W^{(l)})
\end{equation}
where $\tilde{A}=A+I_N$ is the adjacency matrix with added self-connections. $\tilde{D}$ is the diagonal matrix with elements $\tilde{D}_{ii}=\sum_j^N\tilde{A}_{ij}$ and $W^l$ is the trainable parameter matrix. $H^{(l)}$ is the output matrix of the previous layer; $H^{(0)}=X$.

\begin{figure}[!ht]
    \centering
    \includegraphics[width=0.7\textwidth]{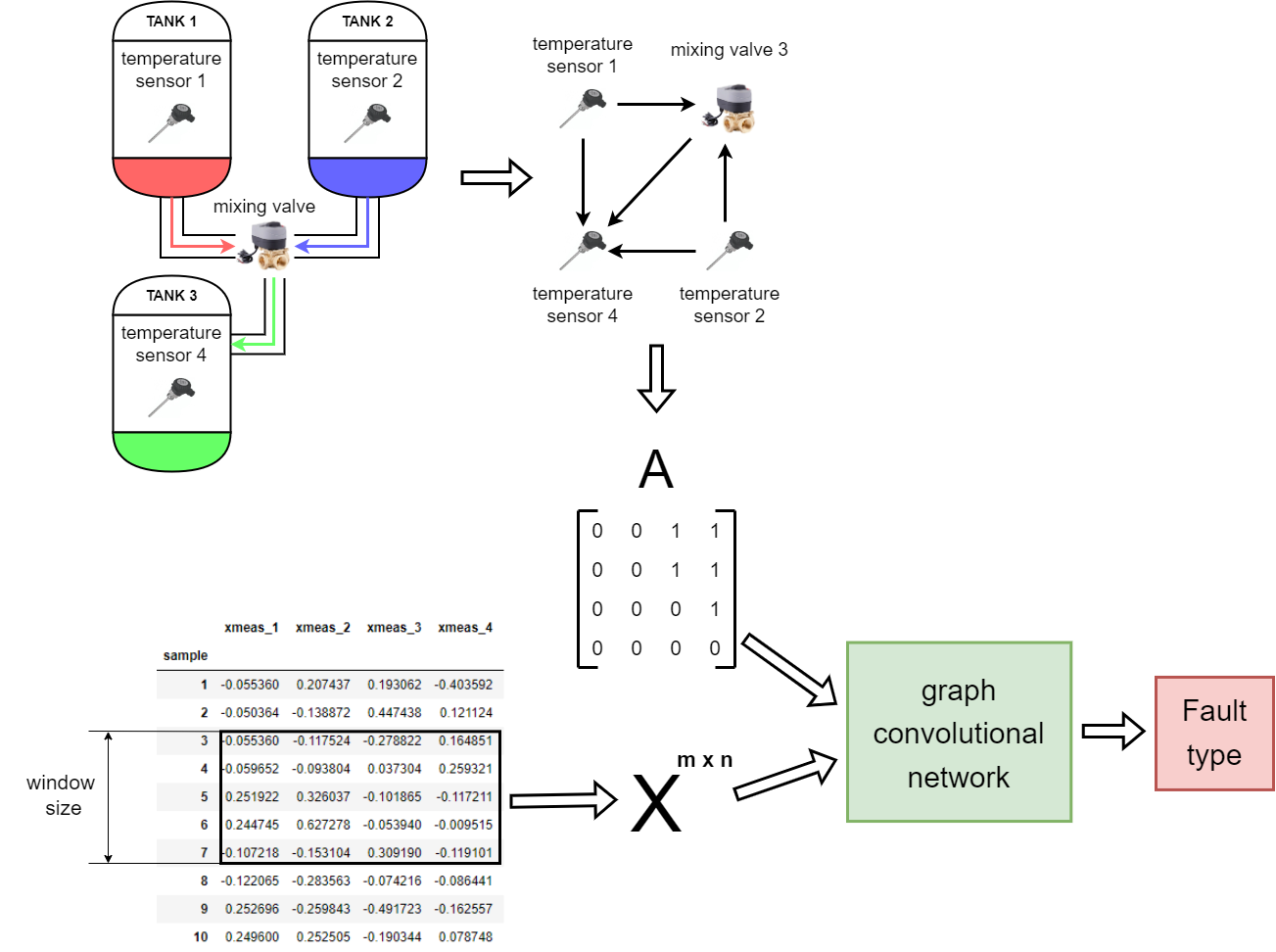}
    \caption{Applying GNN to multivariate sensor data.}
    \label{fig:GNN}
\end{figure}

The diagram of an example of applying GNN to multivariate sensor data is shown in Fig. \ref{fig:GNN}. Production equipment is represented as a graph described by an adjacency matrix. An adjacency matrix is fed into the GCN input together with the feature matrix $X^{m \times n}$, where $m$ is the window size of the time series and $n$ is the number of sensors. The output of GCN is the type of fault.

Several papers on multivariate time series forecasting \cite{GNNForcasting1, GNNForcasting2, GNNForcasting3} using spectral temporal graph neural networks and spatial temporal graph neural networks have been published in recent years. These works solve graph regression problems with benchmark datasets such as traffic, electricity consumption, solar energy generation, and exchange rates. Adjacency matrices were obtained by trainable parameters during the models training process.

Zhiwen Chen et al. \cite{GNNFD} made a review on graph neural network-based fault diagnosis. Four GNN architectures have been designed for fault multi-class classification task: graph convolutional network, graph attention network (GAT), graph sample and aggregate (GraphSage) and spatial temporal graph convolutional network (STGCN). The adjacency matrices were obtained by k nearest neighbors (KNN) and KNN + Graph autoencoder (GAE) methods. KNN approach was applied to the similarity of time-frequency features. GNN-based methods showed better performance than the six baseline machine learning approaches. Tianfu Li et al. \cite{GNNFD2} proposed a practical guideline and a GNN-based FDD framework. In this framework, two types of graph construction methods for multivariate time series where discovered: KNN-based and Radius-based. These methods are applied to the training dataset and the resulting adjacency matrix is fed into the GNN. In this work, the adjacency matrix initially consists of random trainable parameters that are trained together with the entire model.

\section{Model Description}

The adjacency matrix must be known in order to apply graph neural networks. In multivariate time series fault diagnosis problems, such data are very rare. In \cite{GNNFD, GNNFD2}, authors use classical statistical and other modern methods (KNN-based, Radius-based, KNN + graph auto-encoder etc.) to build an adjacency matrix. All these methods are used on the training data and feed the result to the neural network before training starts. In this work, the so-called graph structure learning layer is discovered for the FDD task. This implies that the adjacency matrix is obtained during the training of the GNN and can be adjusted when training on new data.

\begin{figure}[!ht]
    \centering
    \includegraphics[width=1\textwidth]{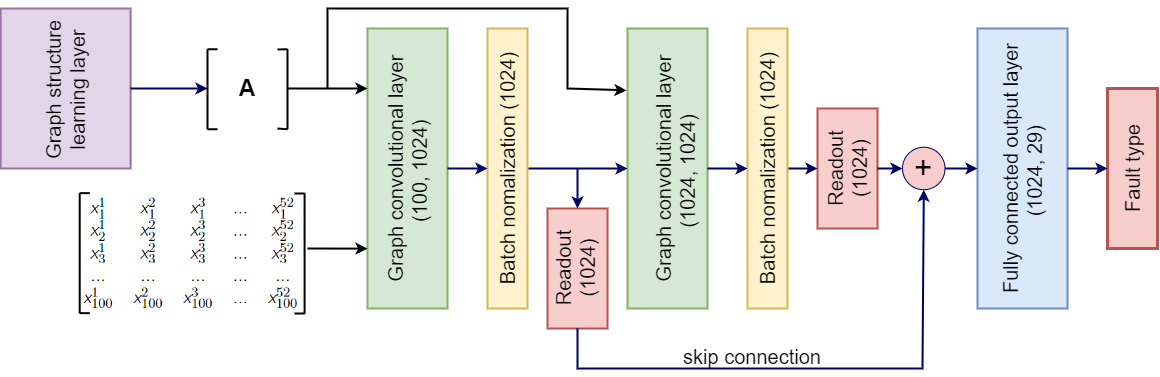}
    \caption{Our model architecture with two GCN layers and graph structure learning layer. Graph structure learning layer builds an adjacency matrix during model training.}
    \label{fig:arc}
\end{figure}

To compare different ways to obtain an adjacency matrix, the general architecture of a graph neural network with two GCN layers is used (Fig. \ref{fig:arc}). There are two read-out layers that create a graph representation by applying a $min$ function to node representations. Batch normalization was added to make the $min$ function more efficiently. Both read-out layers are connected by skip connection. The result of the addition is fed to a fully connected layer that indicates whether the graph belongs to the normal state or to one of the fault classes. Next, motivated by \cite{GNNForcasting3, AM1, AM2, AM3} various options for the graph structure learning layer described.

\begin{itemize}
    \item \textbf{ReLU(W)}. The idea is that the adjacency matrix is a parameter matrix that contains $N^2$ parameters.
    \begin{equation}
    A = \text{ReLU}(W)
    \end{equation}
    The result is a weighted adjacency matrix.
    
    \item \textbf{Uni-directed A}. This approach assumes that the relations between the graph nodes are unidirectional. There cannot be more than one directed edge between two nodes.
    \begin{equation} \label{m1}
    M_1 = \tanh{(\alpha E_1 \Theta_1)}
    \end{equation}
    \begin{equation} \label{m2}
    M_2 = \tanh{(\alpha E_2 \Theta_2)}
    \end{equation}
    \begin{equation} \label{a}
    A = \text{ReLU}(\tanh{(\alpha(M_1 M_2^T - M_2 M_1^T))})
    \end{equation}
    $E_1$ in Eq. \ref{m1} and $E_2$ in Eq. \ref{m2} are randomly initialized trainable node embeddings. $\Theta_1$ in Eq. \ref{m1} and $\Theta_2$ in Eq. \ref{m2} are model parameters and $\alpha$ is a hyperparameter to control the saturation rate of the activation function. Small $\alpha$ values allow us to adjust the order of the weights, while large values bring the weights closer to 1. Expression $(M_1 M_2^T - M_2 M_1^T)$ in Eq. \ref{a} gives asymmetric properties to the resulting adjacency matrix.
    
    \item \textbf{Undirected A}. The undirected graph assumes that the edges between its nodes do not have direction.
    \begin{equation}
    A = \text{ReLU}(\tanh{(\alpha(M_1 M_1^T))})
    \end{equation}
    
    \item \textbf{Directed A}. The directed graph assumes that all its edges have their own directions.
    \begin{equation}
    A = \text{ReLU}(\tanh{(\alpha(M_1 M_2^T))})
    \end{equation}
    
    \item \textbf{Tanh(W)}. The adjacency matrix with both positive and negative weights showed good results during the experiment. The idea is to use hyperbolic tangent as activation function for parameter matrix.
    \begin{equation}
    A = \tanh{(\alpha W)}
    \end{equation}
    
\end{itemize}

It is possible to limit the maximum number of edges in each node and make the adjacency matrix more sparse. This can be useful when the resulting adjacency matrix has too many connections, or when only edges with strong dependency between nodes need to be found. To remove edges with small weights from the graph, the following strategy is proposed. For each node, the \textit{top k} edges with the maximum weights are stored. The rest of the edges weights are set to zero.

\section{Dataset Description}
\begin{figure}[!ht]
    \centering
    \includegraphics[width=1\textwidth]{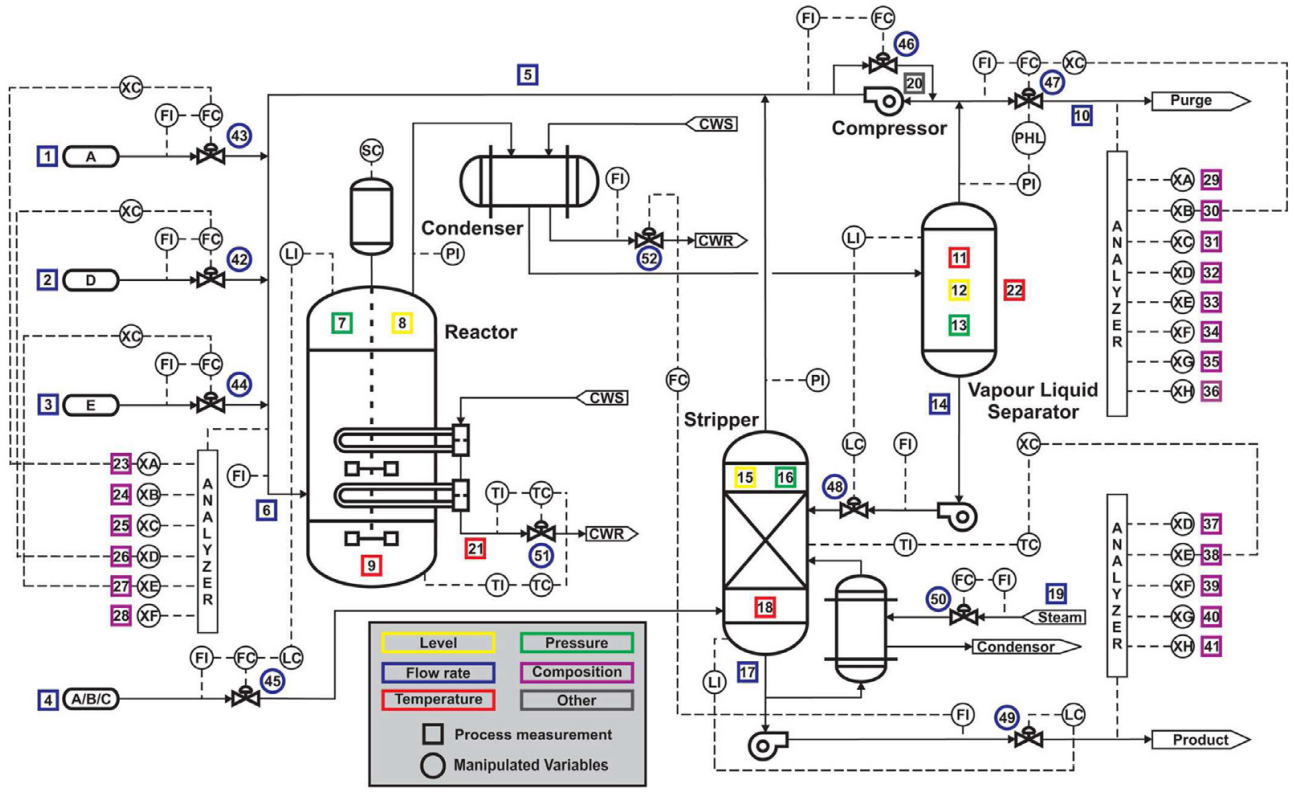}
    \caption{Additionally marked Tennessee Eastman Process diagram (L. Ma, J. Dong and K. Peng \cite{ISA}, 2019).}
    \label{fig:tep1}
\end{figure}

The Tennessee Eastman Process (TEP) dataset is used to evaluate FDD approaches in this work. TEP was created by the Eastman Chemical Company of Tennessee in 1993 \cite{TEP} and became a widely used benchmark in the academic community. This is a flowchart for an industrial plant consisting of five main units: reactor, condenser, stripper, compressor and separator. The process schema is shown in Fig. \ref{fig:tep1}.

Eight components $A$, $B$, $C$, $D$, $E$, $F$, $G$, and $H$ are involved in the chemical process with the following relationships (g -- gaseous, liq -- liquid):
\begin{equation} \label{g}
    A(g) + C(g) + D(g) \rightarrow G(liq)
\end{equation}
\begin{equation} \label{h}
    A(g) + C(g) + E(g) \rightarrow H(liq)
\end{equation}
\begin{equation} \label{f}
    A(g) + E(g) \rightarrow F(liq)
\end{equation}
\begin{equation}
    3D(g) \rightarrow 2F(liq)
\end{equation}

Gaseous reactants $A$, $C$, $D$ and $E$ are involved in the formation of liquid products $G$ and $H$ (Eq. \ref{g} and \ref{h}). $B$ is an inactive component that does not participate in the chemical reaction and $F$ (Eq. \ref{f}) is a byproduct of the process. Gaseous components enter the reactor where the main chemical reaction takes place. Heat released during the reaction is removed by means of a built-in cooling unit. The resulting product and reagent residues enter the condenser for cooling and condensation. Noncondensed components are returned to the reactor through the separator. Inert gases and byproducts are removed from the system as vapor. After the separator, the condensed components enter the stripper, where they are removed from excess reagents and left in the form of products $G$ and $H$.

The TEP dataset consists of multivariate time series simulations referring to either the normal state of the process or one of 28 fault types (Table \ref{table:1}). Each multivariate time series simulation has 41 measured variables and 11 control variables from different sensors recorded with 3-minute intervals. 

Initial TEP dataset contains insufficient number of time series examples for training deep neural networks. The authors of \cite{eTEP} have extended it by generating more examples for each failure/normal state. The additional Tennessee Eastman Process dataset contains 100 simulations for each type of faults. 80 of them were used for training and 20 for test cases. Each simulation contains 2000 data points. The fault states appear after the 600th point.
\begin{table}[!ht]
\caption{Description of the Normal/Fault states in TEP dataset.}
\label{table:1}
\centering
\resizebox{\textwidth}{!}{
\begin{tabular}{ |c|c|c| } 
 \hline
 State ID & Description & Type \\ 
 \hline
 0 & Normal state & None \\ 
 1 & A/C feed ratio, B composition constant & Step \\ 
 2 & B composition, A/C ratio constant & Step \\ 
 3 & D feed temperature & Step \\
 4 & Reactor cooling water inlet temperature & Step \\ 
 5 & Condenser cooling water inlet temperature & Step \\
 6 & A feed loss & Step \\ 
 7 & C header pressure loss-reduced availability & Step \\
 8 &  A, B, C feed composition & Random Variation \\ 
 9 & D feed temperature & Random Variation \\
 10 & C feed temperature & Random Variation \\ 
 11 & Reactor cooling water inlet temperature & Random Variation \\
 12 & Condenser cooling water inlet temperature & Random Variation \\ 
 13 & Reaction kinetics &  Slow Drift \\
 14 & Reactor cooling water valve & Sticking \\ 
 15 & Condenser cooling water valve & Sticking \\
 16 & Unknown & Unknown \\ 
 17 & Unknown & Unknown \\
 18 & Unknown & Unknown \\ 
 19 & Unknown & Unknown \\
 20 & Unknown & Unknown \\
 21 & Unknown & Unknown \\
 22 & Unknown & Unknown \\
 23 & Unknown & Unknown \\
 24 & Unknown & Unknown \\
 25 & Unknown & Unknown \\
 26 & Unknown & Unknown \\
 27 & Unknown & Unknown \\
 28 & Unknown & Unknown \\
 \hline
\end{tabular}}
\end{table}

\section{Experiment}

During the research, the experiment was divided into three parts. At the first step, the efficiency of the GNN-based model for the fault diagnosis task was investigated. Next, the quality of obtained adjacency matrices was studied by using them in the GNN model with another structure. Finally,  another architecture with several graph structure learning layers was proposed.

\subsection{Fault diagnosis}

The proposed GNN-based model was trained with five different variants of graph structure learning layer. Hyper-parameter $\alpha$ was set to 0.1 after comparing different values. The TEP dataset was normalized by the standard score formula. The node features are presented as time series values with a window size equal to 100. Adam with an initial learning rate of 0.001 was chosen as the optimization algorithm. All models were trained for 40 epochs. 3 training iterations were performed for each model, and the average values of the metrics were taken.

\subsubsection{Metrics}

During the evaluation process, the model receives a sample as input and labels it as one of the process states. When comparing the result with labeled test data, four main outcomes are possible:
\begin{itemize}
    \item True positive (TP) is the case where the faulty process state was correctly diagnosed.
    \item False positive (FP) is the case where the normal process state was diagnosed as a fault.
    \item True negative (TN) is the case where the normal process state was correctly diagnosed.
    \item False negative (FN) is the case where the faulty process state was diagnosed as normal one.
\end{itemize}

The following metrics are very important in FDD tasks and were used to evaluate the quality of the models:

\begin{itemize}
    \item[$\blacksquare$] True Positive Rate (TPR) is the probability that the fault state is diagnosed correctly:
    \begin{equation}
    TPR = \frac{TP}{TP+FN}
    \end{equation}
    \item[$\blacksquare$] False Positive Rate (FPR) is the probability that the normal state is detected as a fault: 
    \begin{equation}
    FPR = \frac{FP}{FP+TN}
    \end{equation}
    \item[$\blacksquare$] Average Detection Delay (ADD) is an average time delay between the moment when the state of the process changed (from normal to faulty condition) and the moment when the model detected this change.
\end{itemize}\par
There are two types of tables to compare the models. The first type contains TPRs and FPRs calculated separately for each type of fault. The second type contains detection TPR, detection FPR, and ADD. Detection TPR/FPR is the probability that the faulty state is detected correctly/incorrectly (irrespective of the correct fault classification).

\subsubsection{Baseline models}

The model with a predefined adjacency matrix was taken as a baseline. To obtain the adjacency matrix, the correlation method was used. Based on the dataframe, the Pearson correlation matrix was built. Values less than 0.3 were removed. This adjacency matrix generation strategy performed better in preliminary tests compared to other methods that have been tried.

\begin{figure}[!ht]
    \centering
    \includegraphics[width=0.7\textwidth]{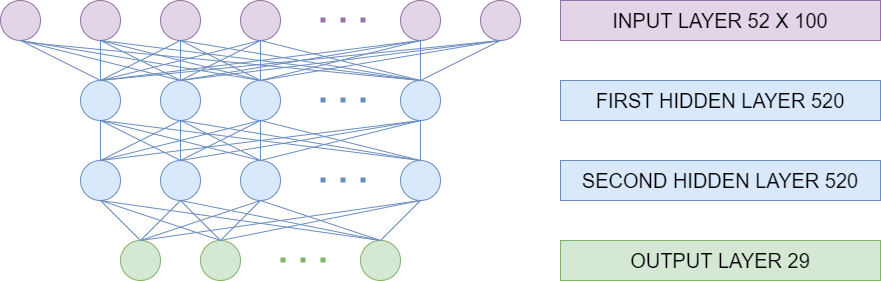}
    \caption{MLP with two hidden layers.}
    \label{fig:MLP}
\end{figure}

To test that GNN-based models can compete with other types of neural networks, MLP, 1DCNN \cite{1dcnn}, and GRU \cite{RNN} based models where also added to the baseline. MLP with two hidden layer (Fig. \ref{fig:MLP}) was chosen as a simple model with cheap computation costs.

 \par

\begin{figure}[!ht]
    \centering
    \includegraphics[width=0.8\textwidth]{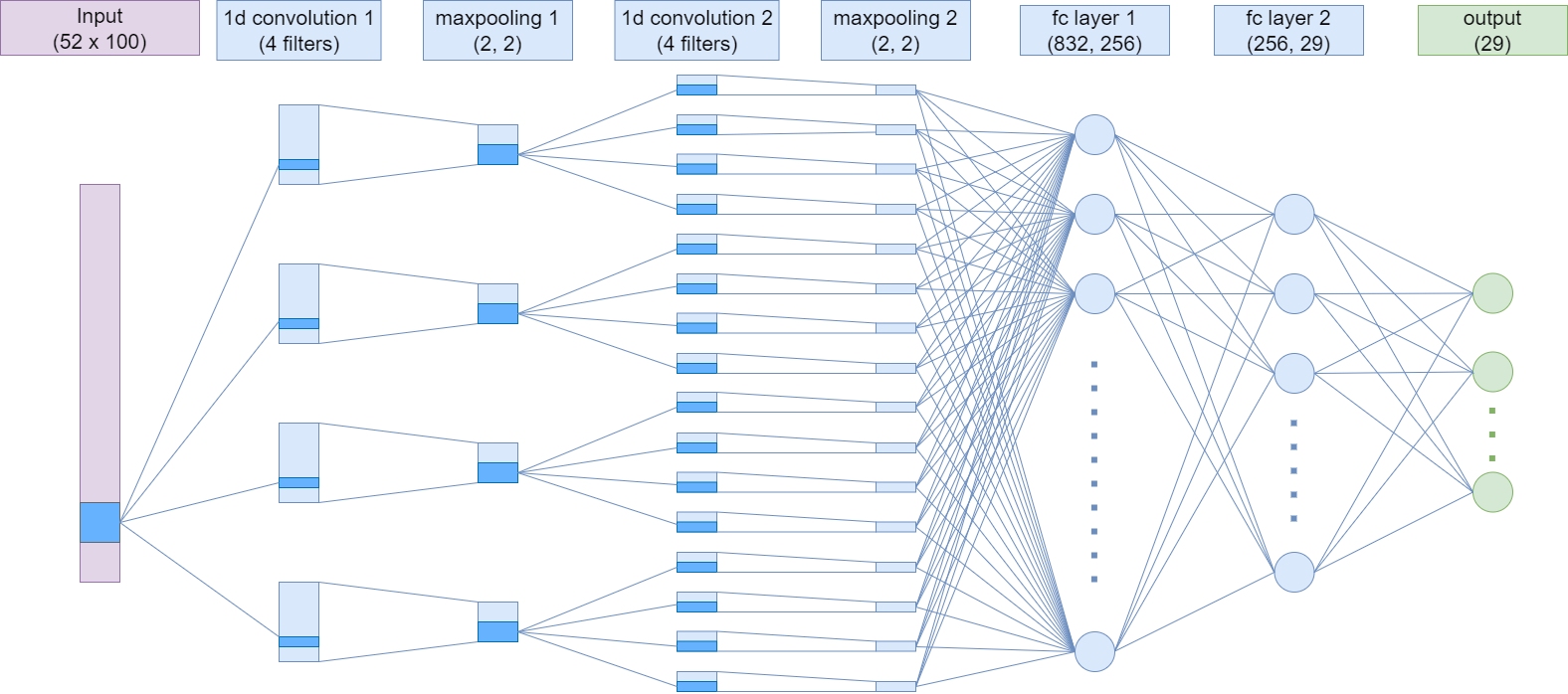}
    \caption{1DCNN baseline architecture.}
    \label{fig:CNN}
\end{figure}

In recent years, convolutional neural networks have shown promising results in fault diagnosis tasks. As a CNN baseline, several standard architectures with one and two 1d convolutional layers were created. The best TPR/FPR scores were obtained by the model with two 1d convolutional and maxpooling layers. Architecture depicted in Fig. \ref{fig:CNN}.

\begin{figure}[!ht]
    \centering
    \includegraphics[width=0.6\textwidth]{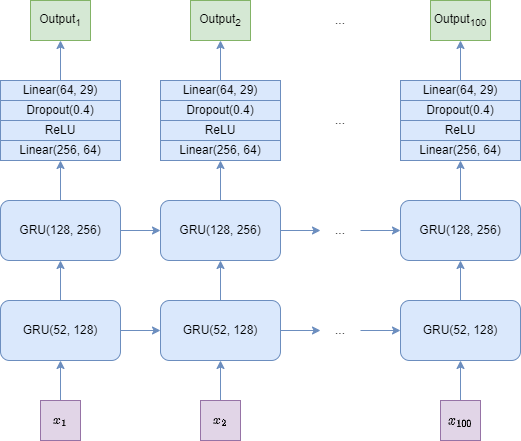}
    \caption{GRU baseline architecture.}
    \label{fig:gru_type2}
\end{figure}

Recurrent Neural Network (RNN) is an another type of deep learning models that show high quality in modeling sequential data. In particular, authors of \cite{RNN} have showed that a network based on Gated Recurrent Units (GRU) can achieve the best results in supervised FDD on TEP dataset. We selected the GRU architecture that referred to "GRU: type 2" in the paper. The schematic view of the model is depicted in Figure \ref{fig:gru_type2}.

\subsubsection{Results}

Detection TPR, Detection FPR, ADD and the number of trainable parameters for each model are shown in Table \ref{table:2}. The best detection TPR scores were obtained by GNN + Tanh(W) model. All models with graph structure learning layers showed TPR / FPR scores similar to those of the GRU model and even outperformed it. The very important ADD metric has been significantly reduced. The experiment confirmed that graph neural networks can compete with other types of NN and show the best results in FDD task on TEP dataset. The detailed results for each fault type are shown in Table \ref{table:3}.

\begin{center}
\begin{table}[!ht]
\footnotesize
\caption{Values of detection TPR/FPR, and ADD for each model. Best results are in bold.}
\label{table:2}
\centering
\resizebox{\textwidth}{!}{
\begin{tabular}{ |c|c|c|c|c|c| } 
 \hline
 & Model & Detection TPR & Detection FPR & ADD & Num. of trainable par. \\ 
 \hline
 Ours & GNN + ReLU(W) & 0.9265 & 0.0317 & 20.67 & 1185661 \\ 
 & GNN + Uni-directed A & 0.9189 & 0.0113 & 31.27 & 1213557  \\ 
 & GNN + Undirected A & 0.9262 & 0.0130 & 29.53 & 1198257  \\ 
 & GNN + Directed A & 0.9301 & 0.0249 & 21.08 & 1213557 \\
 & GNN + Tanh(W) & \textbf{0.9346} & 0.0406 & \textbf{18.95} & 1185661 \\
 \hline
 Baselines & GNN + Correlation & 0.8882 & \textbf{0.0064} & 46.86 & \textbf{99309} \\
 & MLP & 0.8895 & 0.0136 & 28.06 & 2688189  \\ 
 & CNN1D & 0.8967 & 0.0151 & 25.07 & 226941  \\
 & GRU & 0.9247 & 0.0387 & 35.46 & 384669 \\
 \hline
\end{tabular}}
\end{table}
\end{center}

\begin{center}
\begin{table}[!ht]
\footnotesize
\caption{Results of fault diagnosis for each type of fault. Best results are in bold.}
\label{table:3}
\centering
\resizebox{\textwidth}{!}{
\begin{tabular}{|c|c|c|c|c|c|c|c|c|c|} 
 \hline
 Fault & GNN + & GNN + & GNN + & GNN + & GNN + & GNN + & MLP & 1DCNN & GRU\\ 
 ID & ReLU & Uni-dir. & Undir. & Dir. & Tanh & Corr. & & &\\ 
 \hline
 1 & \textbf{1.00}/0.00 & 0.99/0.00 &   \textbf{1.00}/0.00 & \textbf{1.00}/0.00 & \textbf{1.00}/0.00 & \textbf{1.00}/0.00 & \textbf{1.00}/0.00 & \textbf{1.00}/0.00 & \textbf{1.00}/0.00\\
 2 & 0.99/0.00 & 0.99/0.00 &   0.99/0.00 & 0.99/0.00 & 0.99/0.00 & 0.99/0.00 & 0.99/0.00 & 0.99/0.00 & 0.99/0.00\\
 3 & 0.98/0.00 & 0.97/0.00 &   \textbf{0.99}/0.00 & 0.97/0.00 & 0.97/0.00 & 0.95/0.00 & 0.97/0.00 & 0.96/0.00 & 0.98/0.00\\
 4 & \textbf{1.00}/0.00 & \textbf{1.00}/0.00 &   \textbf{1.00}/0.00 & \textbf{1.00}/0.00 & \textbf{1.00}/0.00 & \textbf{1.00}/0.00 & \textbf{1.00}/0.00 & 0.99/0.00 & \textbf{1.00}/0.00\\
 5 & 0.99/0.00 & 0.99/0.00 &   0.99/0.00 & 0.99/0.00 & 0.99/0.00 & \textbf{1.00}/0.00 & 0.98/0.00 & 0.97/0.00 & 0.99/0.00\\
 6 & 1.00/0.00 & 1.00/0.00 &   1.00/0.00 & 1.00/0.00 & 1.00/0.00 & 1.00/0.00 & 1.00/0.00 & 1.00/0.00 & 1.00/0.00\\
 7 & 1.00/0.00 & 1.00/0.00 &   1.00/0.00 & 1.00/0.00 & 1.00/0.00 & 1.00/0.00 & 1.00/0.00 & 1.00/0.00 & 1.00/0.00\\
 8 & 0.98/0.00 & 0.98/0.00 &   0.98/0.00 & 0.98/0.00 & 0.98/0.00 & 0.98/0.00 & 0.98/0.00 & 0.98/0.00 & 0.98/0.00\\
 9 & 0.58/0.00 & 0.62/0.00 &   0.61/0.00 & 0.58/0.00 & 0.58/0.00 & 0.54/0.00 & \textbf{0.70}/0.00 & 0.49/0.00 & 0.57/0.00\\
 10 & 0.97/0.00 & \textbf{0.98}/0.00 &  \textbf{0.98}/0.00 & \textbf{0.98}/0.00 & \textbf{0.98}/0.00 & 0.96/0.00 & 0.97/0.00 & \textbf{0.98}/0.00 & \textbf{0.98}/0.00\\
 11 & \textbf{0.99}/0.00 & \textbf{0.99}/0.00 &  \textbf{0.99}/0.00 & \textbf{0.99}/0.00 & \textbf{0.99}/0.00 & \textbf{0.99}/0.00 & 0.98/0.00 & \textbf{0.99}/0.00 & \textbf{0.99}/0.00\\
 12 & 0.98/0.00 & \textbf{0.99}/0.00 &  \textbf{0.99}/0.00 & \textbf{0.99}/0.00 & 0.98/0.00 & 0.98/0.00 & 0.98/0.00 & 0.98/0.00 & \textbf{0.99}/0.00\\
 13 & 0.97/0.00 & 0.97/0.00 &  0.97/0.00 & 0.97/0.00 & 0.97/0.00 & 0.97/0.00 & 0.97/0.00 & 0.97/0.00 & 0.97/0.00\\
 14 & \textbf{1.00}/0.00 & \textbf{1.00}/0.00 &  \textbf{1.00}/0.00 & \textbf{1.00}/0.00 & \textbf{1.00}/0.00 & \textbf{1.00}/0.00 & 0.99/0.00 & 0.99/0.00 & \textbf{1.00}/0.00\\
 15 & 0.39/0.01 & 0.41/0.01 &  0.39/0.01 & 0.48/0.01 & \textbf{0.49}/0.01 & 0.00/0.00 & 0.01/0.01 & 0.08/0.00 & 0.26/0.02\\
 16 & 0.87/0.00 & 0.84/0.00 &  0.87/0.01 & 0.85/0.00 & \textbf{0.94}/0.00 & 0.65/0.00 & 0.67/0.00 & 0.67/0.00 & 0.90/0.01\\
 17 & 0.98/0.00 & 0.98/0.00 &  0.98/0.00 & 0.98/0.00 & 0.98/0.00 & 0.98/0.00 & 0.98/0.00 & 0.98/0.00 & 0.98/0.00\\
 18 & 0.96/0.00 & 0.96/0.00 &  0.96/0.00 & 0.96/0.00 & 0.96/0.00 & 0.96/0.00 & 0.96/0.00 & 0.96/0.00 & 0.96/0.00\\
 19 & 0.99/0.00 & 0.99/0.00 &  0.99/0.00 & 0.99/0.00 & 0.99/0.00 & 0.99/0.00 & 0.99/0.00 & 0.99/0.00 & 0.99/0.00\\
 20 & 0.97/0.00 & 0.97/0.00 &  0.97/0.00 & 0.97/0.00 & 0.97/0.00 & 0.97/0.00 & 0.97/0.00 & 0.97/0.00 & 0.97/0.00\\
 21 & 0.02/0.01 & 0.00/0.00 &  0.00/0.00 & 0.01/0.00 & \textbf{0.04}/0.02 & 0.00/0.00 & 0.01/0.00 & 0.00/0.00 & 0.02/0.02\\
 22 & 0.60/0.00 & 0.59/0.00 &  \textbf{0.71}/0.00 & 0.64/0.00 & 0.62/0.00 & 0.69/0.00 & 0.47/0.00 & 0.61/0.00 & 0.69/0.00\\
 23 & 0.91/0.00 & 0.84/0.00 &  0.93/0.00 & 0.94/0.00 & \textbf{0.97}/0.00 & 0.71/0.00 & 0.63/0.00 & 0.78/0.00 & 0.95/0.01\\
 24 & 0.99/0.00 & 0.99/0.00 &  0.99/0.00 & 0.99/0.00 & 0.99/0.00 & 0.99/0.00 & 0.99/0.00 & 0.99/0.00 & 0.99/0.00\\
 25 & \textbf{0.99}/0.00 & \textbf{0.99}/0.00 &  \textbf{0.99}/0.00 & \textbf{0.99}/0.00 & \textbf{0.99}/0.00 & \textbf{0.99}/0.00 & 0.98/0.00 & \textbf{0.99}/0.00 & \textbf{0.99}/0.00\\
 26 & \textbf{0.98}/0.00 & \textbf{0.98}/0.00 &  \textbf{0.98}/0.00 & \textbf{0.98}/0.00 & \textbf{0.98}/0.00 & \textbf{0.98}/0.00 & 0.97/0.00 & \textbf{0.98}/0.00 & \textbf{0.98}/0.00\\
 27 & \textbf{0.99}/0.00 & \textbf{0.99}/0.00 &  \textbf{0.99}/0.00 & \textbf{0.99}/0.00 & \textbf{0.99}/0.00 & \textbf{0.99}/0.00 & 0.98/0.00 & \textbf{0.99}/0.00 & \textbf{0.99}/0.00\\
 28 & 0.95/0.00 & 0.95/0.00 &  \textbf{0.96}/0.00 & \textbf{0.96}/0.00 & \textbf{0.96}/0.00 & 0.85/0.00 & 0.84/0.01 & 0.90/0.00 & 0.93/0.01\\
 \hline
\end{tabular}}
\end{table}
\end{center}

All models with graph structure learning layers shown better scores than the baseline with predefined adjacency matrix. The adjacency matrices obtained during the training can be very useful in faster troubleshooting of equipment malfunctions. Their quality is studied in the next section.

\subsection{Quality of obtained adjacency matrices}
\begin{figure}[!ht]
    \centering
    \includegraphics[width=0.7\textwidth]{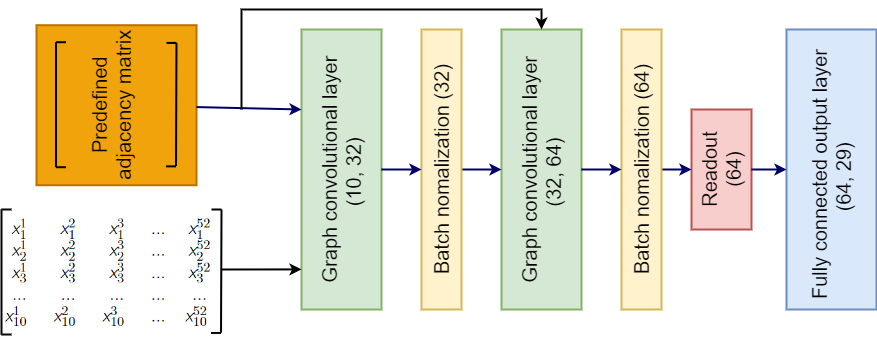}
    \caption{Simplified model architecture with two GCN layers and a predefined adjacency matrix.}
    \label{fig:arc2}
\end{figure}

To obtain an adjacency matrix, the graph structure learning layer parameters are trained together with the parameters of the other layers and can be adapted to their values. On the one hand, the adjacency matrix must correctly indicate the dependencies between the nodes so that subsequent GCN layers correctly process the incoming data. On the other hand, during backpropagation, the trainable weights of the adjacency matrix depend on the weights of the GCN layers. To evaluate this negative influence, a simplified model with a different structure, another set of parameters and window size equals to 10 is proposed (Fig. \ref{fig:arc2}). The adjacency matrices obtained during the fault diagnosis step in the previous subsection are fed into the model as predefined ones, and the results is compared with the baseline (matrix obtained by correlation). The results in Table \ref{table:4} show that the trained adjacency matrices outperform the predefined one in TPR and FPR scores. This may indicate that, despite the influence of the weights of the GCN layers, the trained adjacency matrices can well display the dependencies in the data.

\begin{center}
\begin{table}[!ht]
\footnotesize
\caption{Values of detection TPR/FPR, and ADD for simplified models with two GCN layers and a predefined adjacency matrix. Best results are in bold.}
\label{table:4}
\centering
\resizebox{\textwidth}{!}{
\begin{tabular}{ |c|c|c|c|c|c| } 
 \hline
 & Model & Detection TPR & Detection FPR & ADD & Num. of trainable par. \\ 
 \hline
 Ours & GNN + ReLU(W) & 0.7667 & 0.0008 & 46.76 & 4557\\ 
 & GNN + Uni-directed A & 0.7827 & 0.0049 & \textbf{36.21} & 4557\\ 
 & GNN + Undirected A & 0.7735 & 0.0020 & 41.17 & 4557\\ 
 & GNN + Directed A & \textbf{0.7884} & \textbf{0.0012} & 42.95 & 4557\\
 \hline
 Baseline & GNN + Correlation & 0.7491 & 0.0076 & 39.82 & 4557\\
 \hline
\end{tabular}}
\end{table}
\end{center}

\begin{figure}[!ht]
    \centering
    \includegraphics[width=0.7\textwidth]{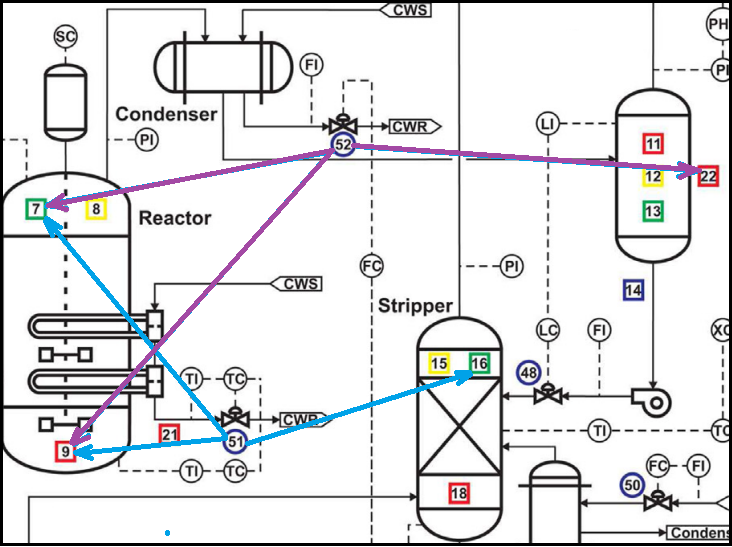}
    \caption{Relationships between sensors received by the graph structure learning layer.}
    \label{fig:tep2}
\end{figure}

When comparing the obtained adjacency matrices and the scheme of the technological process, both correspondences and differences can be found. The differences can be explained by the fact that it is not always possible to see hidden dependencies in the diagram. However, some logical connections can still be found. During the experiment, a model with maximum number of edges for one node equal to 3 was trained. Directed A was chosen as a learning layer. Fig. \ref{fig:tep2} shows the outgoing edges for nodes 51 and 52. Sensor 51 displays the flow rate of the coolant in the reactor, which in turn affects the temperature released during the chemical reaction measured by sensor 9. When the temperature of a substance changes in a closed volume, the pressure changes. Consequently, when the coolant flow changes, not only the temperature changes, but also the pressure recorded by sensor 7. The reactor and stripper tanks are connected, so the pressure also changes on the readings of sensor 16. Sensor 52 measures the coolant flow in the condenser, which also affects sensors 7 and 9 in the reactor. After the condenser, the substance enters the separator, where the temperature of the coolant is measured by sensor 22. The edges obtained using the model are logical and prove that the graph sctructure learning layer can be used to find hidden dependencies between parts of equipment. However, the adjacency matrices obtained at different iterations may differ.

\begin{figure}[!ht]
    \centering
    \includegraphics[width=1\textwidth]{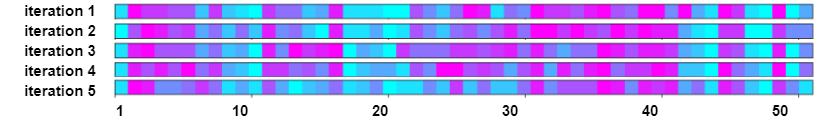}
    \caption{The sums of edge weights belonging to numbered sensors/nodes for different 5 training iterations.}
    \label{fig:matrices}
\end{figure}

The previous model with limited edges was trained iteratively 5 times. All obtained adjacency matrices have common features (some common nodes have large weights, some less). Similarities are shown in Fig. \ref{fig:matrices}. The rows contain the sums of edge weights belonging to the numbered sensors/nodes. It can be seen from the picture that all adjacency matrices have similar node importance, indicating that some of the sensors have a greater influence on the entire system, and some less. This also confirms that the graph structure learning layers does indeed learn the interaction structure in the workflow. The resulting adjacency matrices are not similar to the adjacency matrix obtained by the simple correlation method. Therefore, they can display hidden dependencies between sensors. Such data can be very useful for engineers working with this equipment. In further research, methods for stabilizing the results at different iterations of model training can be studied.

The differences between obtained adjacency matrices also led to the idea that there cannot be an ideal adjacency matrix. The processes of such a complex technological system are not static and can change over time. Depending on the stage of the process, the dependencies between sensors may change. In the next step, the possibility of learning several adjacency matrices in parallel was studied.

\subsection{Model with several graph structure learning layers}

\begin{figure}[!ht]
    \centering
    \includegraphics[width=1\textwidth]{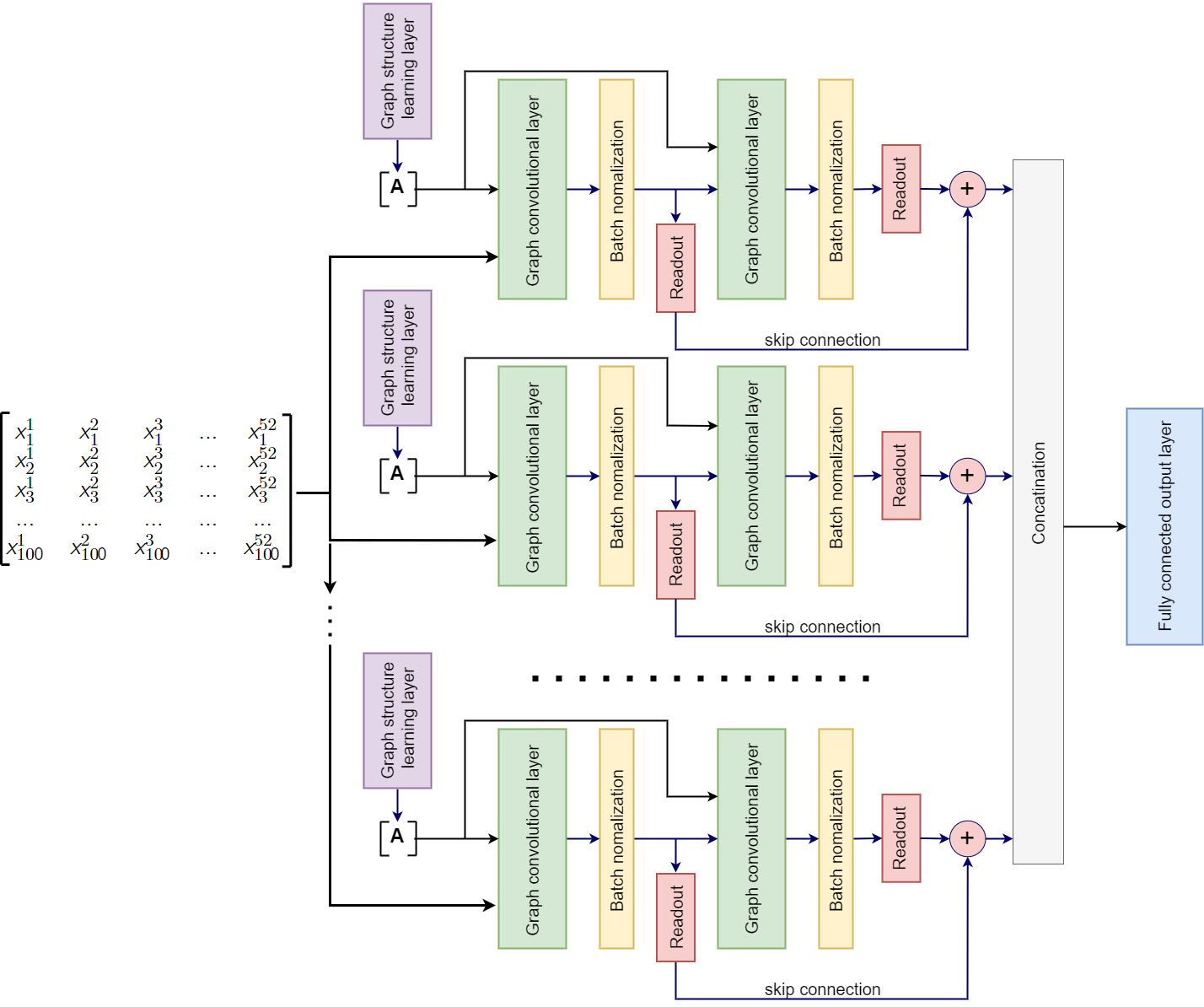}
    \caption{Our model architecture with several GNN modules.}
    \label{fig:arc3}
\end{figure}

To train several adjacency matrices, an architecture with several trained in parallel graph neural networks (GNN modules) is proposed (Fig. \ref{fig:arc3}). The model described in the Model section was taken as an instance of the GNN module. Module outputs are combined by concatenation and fed to the fully connected output layer. Variant with 10 GNN modules was tested. Tanh(W) was chosen as the graph structure learning layer. The number of hidden parameters in graph convolution layers was reduced from 1024 to 64. Thus, the total number of trainable parameters in the proposed architecture is much less than in the original model (153,949 versus 1,185,661). The new model was trained during 35 epochs. The results are shown in Tables \ref{table:5} and \ref{table:6}. 

 \par

\begin{center}
\begin{table}[!ht]
\footnotesize
\caption{Values of detection TPR/FPR, and ADD for a single GNN module and 10 GNN modules. Best results are in bold.}
\label{table:5}
\centering
\resizebox{\textwidth}{!}{
\begin{tabular}{ |c|c|c|c|c| } 
 \hline
 Model & Detection TPR & Detection FPR & ADD & Num. of trainable par. \\ 
 \hline
 GNN + Tanh(W) & 0.9346 & \textbf{0.0406} & \textbf{18.95} & 1185661 \\
 10 x (GNN(64) + Tanh(W)) & \textbf{0.9374} & 0.0533 & 20.79 & \textbf{153949} \\ 
 \hline
\end{tabular}}
\end{table}
\end{center}

\begin{center}
\begin{table}[!ht]
\footnotesize
\caption{Results of fault diagnosis for a single GNN module and 10 GNN modules. Best results are in bold.}
\label{table:6}
\centering
\begin{tabular}{ |c|c|c|c|c|c|c|c| } 
 \hline
 Fault & 1 x GNN(1024) + & 10 x GNN(64) + \\ 
 ID & Tanh(W) & Tanh(W) \\ 
 \hline
 1 & 1.00/0.00 & 1.00/0.00\\
 2 & 0.99/0.00 & 0.99/0.00\\
 3 & 0.97/0.00 & \textbf{0.99}/0.00\\
 4 & 1.00/0.00 & 1.00/0.00\\
 5 & 0.99/0.00 & 0.99/0.00\\
 6 & 1.00/0.00 & 1.00/0.00\\
 7 & 1.00/0.00 & 1.00/0.00\\
 8 & 0.98/0.00 & 0.98/0.00\\
 9 & 0.58/0.00 & \textbf{0.66}/0.00\\
 10 & 0.98/0.00 & 0.98/0.00\\
 11 & 0.99/0.00 & 0.99/0.00\\
 12 & 0.98/0.00 & 0.98/0.00\\
 13 & 0.97/0.00 & 0.97/0.00\\
 14 & 1.00/0.00 & 1.00/0.00\\
 15 & \textbf{0.49}/0.01 & 0.48/0.03\\
 16 & 0.94/0.00 & \textbf{0.95}/0.00\\
 17 & 0.98/0.00 & 0.98/0.00\\
 18 & 0.96/0.00 & \textbf{0.97}/0.00\\
 19 & 0.99/0.00 & 0.99/0.00\\
 20 & 0.97/0.00 & 0.97/0.00\\
 21 & \textbf{0.04}/0.02 & 0.03/0.02\\
 22 & \textbf{0.62}/0.00 & 0.59/0.00\\
 23 & 0.97/0.00 & \textbf{0.99}/0.00\\
 24 & 0.99/0.00 & 0.99/0.00\\
 25 & 0.99/0.00 & 0.99/0.00\\
 26 & 0.98/0.00 & 0.98/0.00\\
 27 & 0.99/0.00 & 0.99/0.00\\
 28 & 0.96/0.00 & 0.96/0.00\\
 \hline
\end{tabular}
\end{table}
\end{center}

Having a significantly smaller number of learning parameters, the proposed architecture shows similar results on the dataset used. At the next stage of the experiment, the training dataset was significantly reduced. Only 10 \% of the simulations for each type of fault were left. Each model was iteratively trained 10 times and the metrics were averaged. The results are shown in Tables \ref{table:7} and \ref{table:8}. On the reduced dataset, the proposed architecture is superior in more fault types than on the full dataset (Tables \ref{table:6} and \ref{table:8}).

\begin{center}
\begin{table}[!ht]
\footnotesize
\caption{Values of detection TPR/FPR, and ADD for a single GNN module and 10 GNN modules on reduced TEP dataset. Best results are in bold.}
\label{table:7}
\centering
\resizebox{\textwidth}{!}{
\begin{tabular}{ |c|c|c|c|c| } 
 \hline
 Model & Detection TPR & Detection FPR & ADD & Num. of trainable par. \\ 
 \hline
 GNN + Tanh(W) & 0.8666 & \textbf{0.0459} & \textbf{29.93} & 1185661 \\ 
 10 x (GNN(64) + Tanh(W)) & \textbf{0.8732} & 0.0547 & 30.49 & \textbf{153949} \\ 
 \hline
\end{tabular}}
\end{table}
\end{center}

\begin{center}
\begin{table}[!ht]
\footnotesize
\caption{Results of fault diagnosis for a single GNN module and 10 GNN modules on reduced TEP dataset. Best results are in bold.}
\label{table:8}
\centering
\begin{tabular}{ |c|c|c|c|c|c|c|c|c| } 
 \hline
 Fault & 1 x GNN(1024) + & 10 x GNN(64) + \\ 
 ID & Tanh(W) & Tanh(W) \\ 
 \hline
 1 & 0.99/0.00 & 0.99/0.000\\
 2 & 0.99/0.00 & 0.99/0.00\\
 3 & \textbf{0.92}/0.00 & 0.91/0.00\\
 4 & 1.00/0.00 & 1.00/0.00\\
 5 & 0.96/0.00 & \textbf{0.97}/0.00\\
 6 & 0.99/0.00 & \textbf{1.00}/0.00\\
 7 & 1.00/0.00 & 1.00/0.00\\
 8 & 0.91/0.00 & \textbf{0.93}/0.00\\
 9 & 0.46/0.00 & \textbf{0.54}/0.00\\
 10 & 0.96/0.00 & \textbf{0.97}/0.00\\
 11 & 0.97/0.00 & \textbf{0.98}/0.00\\
 12 & 0.97/0.00 & \textbf{0.98}/0.00\\
 13 & 0.93/0.00 & \textbf{0.96}/0.00\\
 14 & 1.00/0.00 & 1.00/0.00\\
 15 & \textbf{0.11}/0.01 & 0.04/0.01\\
 16 & 0.62/0.01 & \textbf{0.63}/0.01\\
 17 & 0.97/0.00 & 0.97/0.00\\
 18 & 0.95/0.00 & \textbf{0.96}/0.00\\
 19 & 0.95/0.00 & \textbf{0.98}/0.00\\
 20 & 0.96/0.00 & 0.96/0.00\\
 21 & 0.01/0.00 & 0.01/0.01\\
 22 & \textbf{0.47}/0.00 & 0.40/0.00\\
 23 & 0.02/0.01 & \textbf{0.15}/0.02\\
 24 & 0.97/0.00 & \textbf{0.98}/0.00\\
 25 & 0.98/0.00 & 0.98/0.00\\
 26 & 0.96/0.00 & \textbf{0.97}/0.00\\
 27 & \textbf{0.99}/0.00 & 0.98/0.00\\
 28 & \textbf{0.89}/0.00 & 0.87/0.00\\
 \hline
\end{tabular}
\end{table}
\end{center}

This part of the experiment show that the architecture with several parallel GNN modules, having a much smaller number of parameters, is better at finding dependencies in the data. Especially on reduced TEP dataset. 

\clearpage

\section{Conclusion}

GNN-based models outperform other types of neural networks in the FDD task for the TEP dataset. The structure of the graph, including the relationship between nodes, is not always known. The proposed graph structure learning layer allows us to solve this problem. The adjacency matrices obtained during the training showed a better result than those obtained by the correlation method. The possibility of obtaining hidden relationships between equipment sensors can open up wide opportunities in process control, as well as speed up the repair and adjustment of equipment. In further works, models with graph structure learning layers can be tested on datasets where adjacency matrices are known in advance. Other types of GNN layers can also be considered.

The adjacency matrices obtained during training have similar characteristics (the same nodes have a greater influence on other parts of the equipment, others less), but there are also noticeable differences. This can be explained both by the fact that the influences between nodes can be expressed differently through intermediate nodes and by the fact that the equipment can operate in different modes and the dependencies are not static. During the research, the idea of learning multiple adjacency matrices came up. A model consisting of several parallel GNN modules with graph structure learning layers was proposed. This architecture has shown that, having significantly smaller number of trainable parameters, it shows similar results. It can also show better performance on small datasets.

Graph neural networks have shown good results and prospects in processing data from various types of sensors in technological equipment. Further research may be aimed not only at improving the accuracy of the classification of fault types, but also at ways of interpreting the results of the models. The ability to obtain adjacency matrices can open up great possibilities in determining the root cause of a malfunction, as well as the fault propagation path.

\bibliographystyle{elsarticle-num} 
\bibliography{cas-refs}

\end{document}